\documentclass[11pt]{article}
\usepackage{acl2017}
\usepackage{times}
\usepackage{latexsym}
\usepackage{multirow}
\usepackage{caption}
\usepackage{subcaption}
\usepackage{microtype}
\usepackage{url}
\usepackage{latexsym}
\usepackage{graphicx}
\usepackage{fixltx2e}
\usepackage{epstopdf}
\usepackage{amssymb}
\usepackage{amsmath}
\usepackage{xspace,relsize,gensymb}
\usepackage{graphics,boxedminipage}
\usepackage{adjustbox,paralist}
\usepackage{booktabs}
\usepackage{arydshln}
\usepackage{soul}
\usepackage{color}

\newcommand{\dataset}[1]{\textsc{#1}\xspace}
\newcommand{\apnews}{\dataset{apnews}}

\newcommand{\bnc}{\dataset{bnc}}

\newcommand{\method}[1]{\texttt{#1}\xspace}
\newcommand{\lda}{\method{lda}}
\newcommand{\ctm}{\method{ctm}}
\newcommand{\hca}{\method{hca}}
\newcommand{\ntm}{\method{ntm}}
\newcommand{\cluster}{\method{cluster}}
\newcommand{\wordtovec}{\method{word2vec}}
\newcommand{\pmi}{\method{pmi}}
\newcommand{\npmi}{\method{npmi}}

\newcommand{\ex}[1]{\textit{#1}\xspace}

\newcommand{\secref}[2][]{Section#1~\ref{sec:#2}}
\newcommand{\figref}[2][]{Figure#1~\ref{fig:#2}}
\newcommand{\tabref}[2][]{Table#1~\ref{tab:#2}}

\newcommand\email{\begingroup \urlstyle{tt}\smaller\Url}

\newcommand{\tl}[2]{\multirow{#1}{*}{\begin{tabular}[l]{@{}l@{}l}#2\end{tabular}}}

\DeclareRobustCommand{\hla}[1]{{\sethlcolor{pink}\hl{#1}}}
\DeclareRobustCommand{\hlb}[1]{{\sethlcolor{yellow}\hl{#1}}}

\aclfinalcopy 
\def\aclpaperid{258}

\title{An Automatic Approach for Document-level Topic Model Evaluation}

\author{Shraey Bhatia$^{1}$ \qquad Jey Han Lau$^{1,2}$ \qquad Timothy Baldwin$^{1}$ \\[1ex]
    $^1$ School of Computing and Information Systems,\\The University of
Melbourne \\[0.5ex]
    $^2$ IBM Research \\[1ex]
    \email{shraeybhatia@gmail.com}, \email{jeyhan.lau@gmail.com}, \email{tb@ldwin.net}}
  \date{}

\begin{document}

\maketitle

\begin{abstract}
  Topic models jointly learn topics and document-level topic
  distribution.  Extrinsic evaluation of topic models tends to focus
  exclusively on topic-level evaluation, e.g.\ by assessing the
  coherence of topics. We demonstrate that there can be large
  discrepancies between topic- and document-level model quality, and
  that basing model evaluation on topic-level analysis can be
  highly misleading.  We propose a method for automatically predicting
  topic model quality based on analysis of document-level topic
  allocations, and provide empirical evidence for its robustness.
\end{abstract}

\section{Introduction}
\label{sec:intro}

Topic models such as latent Dirichlet allocation \cite{blei+:2003}
jointly learn latent topics (in the form of multinomial distributions
over words) and topic allocations to individual documents (in the form
of multinomial distributions over topics), and provide a powerful means
of document collection navigation and visualisation
\cite{Newman+:2009a,Chaney:Blei:2012,Smith+:2017}. One property of
LDA-style topic models that has contributed to their popularity is that
they are highly configurable, and can be structured to capture a myriad
of statistical dependencies, such as between topics \cite{blei+:2006},
between documents associated with the same individual
\cite{Rosen-Zvi+:2004}, or between documents associated with individuals
in different network relations \cite{Wang:Blei:2011}. This has led to a
wealth of topic models of different types, and the need for methods to
evaluate different styles of topic model over the same document
collections. Test data perplexity is the obvious solution, but it has
been shown to correlate poorly with direct human assessment of topic
model quality \cite{Chang+:2009}, motivating the need for automatic
topic model evaluation methods which emulate human assessment. Research
in this vein has focused primarily on evaluating the quality of
individual topics
\cite{newman+:2010,Mimno+:2011,Aletras+:2013,lau+:2014,Fang+:2016} and
largely ignored evaluation of topic allocations to individual documents,
and it has become widely accepted that topic-level evaluation is a
reliable indicator of the intrinsic quality of the overall topic model
\cite{lau+:2014}. We challenge this assumption, and demonstrate that
topic model evaluation should operate at both the topic and document
levels.

Our primary contributions are as follows: (1) we empirically demonstrate
that there can be large discrepancies between topic- and document-level
topic model evaluation; (2) we demonstrate that previously-proposed
document-level evaluation approaches can be misleading, and propose an
alternative evaluation method; and (3) we propose an automatic approach
to topic model evaluation based on analysis of document-level topic
distributions, which we show to correlate strongly with manual
annotations.


\section{Related Work}
\label{sec:related-work}

Perplexity or held-out likelihood has long been used as an intrinsic
metric to evaluate topic models
\cite{Wallach+:2009}. \newcite{Chang+:2009} proposed two human judgement
tasks, at the topic and document levels, and showed that there is low
correlation between perplexity and direct human evaluations of topic
model quality. The two tasks took the form of ``intruder'' tasks,
whereby subjects were tasked with identifying an intruder topic word for
a given topic, or an intruder topic for a given document. Specifically,
in the word intrusion task, an intruder word was added to the top-5 topic
words, and annotators were asked to identify the intruder
word. Similarly in the topic intrusion task, a document and 4 topics
were presented --- the top-3 topics corresponding to the document and a
random intruder topic --- and subjects were asked to spot the intruder
topic. The intuition behind both methods is that the higher the quality
of the topic or topic allocation for a given document, the easier it
should be to detect the intruder.

\newcite{newman+:2010} proposed to measure topic coherence directly in the form of ``observed coherence'', in which human judges rated topics directly on an ordinal 3-point scale. They experimented with a range of different methods to automate the rating task, and reported the best results by simply aggregating pointwise mutual information (\pmi) scores for different pairings of topic words, based on a sliding window over English Wikipedia.

Building on the work of \newcite{Chang+:2009}, \newcite{lau+:2014} 
proposed an improved method for estimating observed coherence based on 
normalised \pmi (\npmi), and further automated the word intruder 
detection task based on a combination of word association features 
(\pmi, \npmi, CP1, and CP2) in a learn-to-rank model 
\cite{joachims:2006}. Additionally, the authors showed a strong 
correlation between word intrusion and observed coherence, and suggested 
that it is possible to perform topic model evaluation based on 
aggregation of word intrusion or observed coherence scores across all 
topics.

\section{Datasets and Topic Models}
\label{sec:datasets}

We use two document collections for our experiments: \apnews and the
British National Corpus (``\bnc'': \newcite{Burnard:1995}).  
\apnews is a collection of Associated 
Press\footnote{\url{https://www.ap.org/en-gb/}} news articles from 2009 
to 2016, while \bnc is an amalgamation of extracts from different 
sources such as books, journals, letters, and pamphlets. We sample 50K and 15K 
documents from \apnews and \bnc, respectively, to create two datasets for 
our experiments.

In terms of preprocessing, we use Stanford CoreNLP \cite{Manning+:2014}
to tokenise words and sentences. We additionally remove stop
words,\footnote{We use Mallet's stop word list:
  \url{https://github.com/mimno/Mallet/tree/master/stoplists}}
lower-case all word tokens, filter word types which occur less than 10
times, and exclude the top 0.1\% most frequent word types.  Statistics
for each of the preprocessed datasets are provided in
\tabref{dataset-stats}.

Similarly to \newcite{Chang+:2009}, we base our analysis on a
representative selection of topic models, each of which we train over
\apnews and \bnc to generate 100 topics:
\begin{compactitem}
\item \textbf{\lda} \cite{blei+:2003} uses a symmetric Dirichlet prior
  to model both document-level topic mixtures and topic-level word
  mixtures. It is one of the most commonly used topic model
  implementations and serve as a benchmark for comparison. We use
  Mallet's implementation of \lda for our experiments. Note that Mallet
  implements various enhancements to the basic LDA model, including the
  use of an asymmetric--symmetric prior.

\item \textbf{\ctm} \cite{blei+:2006} is an extension of \lda that uses a
  logistic normal prior over topic proportions instead of a Dirichlet
  prior to model correlations between different topics and reduce
  overlap in topic content.

\item \textbf{\hca} \cite{buntine+:2014} is an extension to LDA to capture
  word burstiness \cite{doyle+:2009}, based on the observation that
  there tends to be higher likelihood of generating a word which has
  already been seen recently. Word generation is modelled by a
  Pitman--Yor process \cite{chen+:2011}.

\item   \textbf{\ntm} \cite{cao+:2015} is a neural topic model, where
  topic--word multinomials are modelled as a look-up layer of words, and
  topic--document multinomials are modelled as a look-up layer of
  documents. The output layer of the network is given by the dot product
  of the two vectors. There are 2 variants of \ntm: unsupervised and
  supervised.  We use only the unsupervised variant in our experiments.

\item   \textbf{\cluster} is a baseline topic model, specifically designed to
  produce highly coherent topics but ``bland'' topic allocations. We
  represent word types in the documents with pre-trained \wordtovec
  vectors \cite{Mikolov+:2013b,Mikolov+:2013c}, pre-trained on Google
  News,\footnote{Available
    from: \url{https://code.google.com/archive/word2vec.}}
  and create word clusters using $k$-means clustering ($k=100$) to
  generate the topics. We derive the multinomial distribution for each
  topic based on the cosine distance to the cluster centroid, and linear
  normalisation across all words.




  To generate the topic allocation for a given document, we first
  calculate a document representation based on the mean of the
  \wordtovec vectors of its content words. For each cluster, we 
represent them by calculating the mean \wordtovec vectors of its top-10 
words. Given the document vector and clusters/topics, we calculate the
  similarity of the document to each cluster based on cosine similarity,
  and finally (linearly) normalise the similarities to generate a
  probability distribution.
\end{compactitem}



\begin{table}[t]
\centering
\begin{tabular}{ccc}
\toprule
\textbf{Dataset} & \textbf{\#Docs} & \textbf{\#Tokens} \\
\midrule
\apnews     & 50K    & 15M      \\
\bnc        & 15K    & 18M      \\
\bottomrule
\end{tabular}
\caption{Statistics for the two document collections used in our experiments}
\label{tab:dataset-stats}
\end{table}

\section{Topic-level Evaluation: Topic Coherence}
\label{sec:topic-coherence}

\begin{table}[t]
\centering
\begin{tabular}{ccc}
\toprule
{\textbf{Model}} &  \multicolumn{1}{c}{\textbf{\apnews}}          & \multicolumn{1}{c}{\textbf{\bnc}}
 \\
 \midrule
\lda                          & 0.16            & 0.14         \\
\ctm                          & 0.07            & 0.09         \\
\hca                          & 0.14            & 0.08         \\
\ntm                          & 0.10            & 0.08         \\
\cluster                      & 0.18            & 0.17         \\
\bottomrule
\end{tabular}
\caption{Topic coherence scores (\npmi)}
\label{tab:npmi-results}
\end{table}

\begin{table*}[t]
\begin{center}
\begin{adjustbox}{max width=1.0\textwidth}
\begin{tabular}{cp{14.0cm}}
\toprule
\textbf{Model} & \textbf{Topics} \\
\midrule
\multirow{3}{*}{\lda} &
oil gas drilling gulf spill natural pipeline wells industry energy \\
& computer video screen program text disk windows electronic machine
graphics \\
&health care hospital services medical staff patients service child 
authority\\
\midrule
\multirow{3}{*}{\cluster} &
river creek lake rivers dam tributary lakes reservoir tributaries creeks 
\\
& prohibited forbid prohibiting prohibits violated prohibit contravened 
forbids violate barred\\
& terrace courtyard staircase staircases courtyards walls pergola 
walkway stairways walkways \\
\bottomrule
\end{tabular}
\end{adjustbox}
\end{center}
\caption{Example \lda and \cluster topics.}
\label{tab:ex-topics}
\end{table*}

Pointwise mutual information (and its normalised variant \npmi) is a
common association measure to estimate topic coherence
\cite{newman+:2010,Mimno+:2011,Aletras+:2013,lau+:2014,Fang+:2016}.
Although the method is successful in assessing topic quality, it tells
us little about the association between documents and topics. As we will
see, a topic model can produce topics that are coherent --- in terms of
\npmi association --- but poor descriptor of the overall concepts in the
document collection.

We first compute topic coherence for all 5 topic models over \apnews and 
\bnc using \npmi \cite{lau+:2014} and present the results in 
\tabref{npmi-results}.\footnote{We use the following open source toolkit 
to compute topic coherence: 
\url{https://github.com/jhlau/topic_interpretability}.}
We see that \lda and \cluster perform consistently well
across both datasets. \hca performs well over \apnews but poorly over \bnc.  
Both \ctm and \ntm topics appear to have low coherence over the two 
datasets.


Based on these results, one would conclude that \cluster is a good topic
model, as it produces very coherent topics. To better understand the
nature and quality of the topics, we present a random sample of \lda and
\cluster topics in \tabref{ex-topics}.

Looking at the topics, we see that \cluster tends to include different
inflectional forms of the same word (e.g.\ \ex{prohibited},
\ex{probihiting}) and near-synonyms/sister words (e.g.\ \ex{river},
\ex{lake}, \ex{creeks}) in a single topic. This explains the strong
\npmi association of the \cluster topics.  On the other hand, \lda
discovers related words that collectively describe concepts rather than
just clustering (near) synonyms. This suggests that the topic coherence
metric alone may not completely capture topic model quality, leading us
to also investigate the topic distribution associated with documents
from our collections.

\section{Human Evaluation of Document-level Topic Allocations}
\label{sec:human-judge}

In this section, we describe a series of manual evaluations of
document-level topic allocations, in order to get a more holistic
evaluation of the true quality of the different topic models (in line with
the original work of \newcite{Chang+:2009}).

\subsection{Topic Intrusion}
\label{sec:topic-intrusion}

The goal of the topic intrusion task is to examine whether the 
document--topic allocations from a given topic model accord with manual 
judgements. We formulate the task similarly to \newcite{Chang+:2009}, in 
presenting the human judges with a snippet from each document, along 
with four topics. The four topics comprise the top-3 highest probability 
topics related to document, and one intruder topic. Each annotator is 
required to pick the topic that is least representative of the document, 
with the expectation that the better the topic model, the more readily 
they should be able to pick the intruder topic. The intruder topic is 
sampled randomly, subject to the following conditions: (1) it should be 
a low probability topic for the target document; and (2) it should be a 
high probability topic for at least one other document. 
The first constraint is intended to ensure that the intruder topic is 
unrelated to the target document, while the second constraint is 
intended to select a topic that is highly associated with some 
documents, and hence likely to be coherent and not a junk topic. Each 
topic is represented by its top-10 most probable words, and the target 
document is presented in the form of the first three sentences, with an 
option to view more of the document if further context is needed.

We used Amazon Mechanical Turk to collect the human judgements, with five document--topic combinations forming a single HIT, one of which acts as a quality control. 
The control items were sourced from an earlier annotation task where
subjects were asked to score the top-5 topics for a target document on a
scale of 0--3. The 50 top-scoring documents from this annotation task,
with their top-3 topics, were chosen as controls. The intruder topic for
the control was generated by randomly selecting 10 words from the corpus
vocabulary. In order to pass quality control, each worker had to correctly
select the intruder topic for the control document--topic item over 60\%
of time (across all HITs they completed). Each document--topic pair was
rated by 10 annotators initially, and for HITs where less than 3
annotations passed quality control, we reposted them for a second round
of annotation.

\begin{table}[t]
\centering

\begin{tabular}{ccc}
  \toprule
\multirow{2}{*}{Topic Model} & \multicolumn{2}{c}{Mean Model Precision} \\
  \cmidrule{2-3}
                             & \multicolumn{1}{c}{\apnews}              & \multicolumn{1}{c}{\bnc}              \\
 \midrule
\lda                          & 0.84                & 0.66             \\
\ctm                          & 0.64                & 0.66             \\
\hca                          & 0.60                & 0.44             \\
\ntm                          & 0.26                & 0.17             \\
\cluster                      & 0.39                & 0.48            \\
\bottomrule
\end{tabular}
\caption{Mean model precision for human judgements}
\label{tab:mean-model-precision}
\end{table}

\begin{table}[t]
\centering
\begin{tabular}{ccc}
  \toprule
\multirow{2}{*}{Topic Model} & \multicolumn{2}{c}{Mean Topic Log Odds} \\
  \cmidrule{2-3}
                             & \multicolumn{1}{c}{\apnews}              & \multicolumn{1}{c}{\bnc}               \\
  \midrule
\lda                          & -0.78               & -1.84             \\
\ctm                          & -1.04               & -1.60             \\
\hca                          & -2.09               & -3.61             \\
\ntm                          & -7.16               & -6.32             \\
\cluster                      & -0.12               & -0.10            \\
\bottomrule
\end{tabular}
\caption{Mean topic log odds for human judgements}
\label{tab:mean-tlo}
\end{table}

\begin{figure*}[t]
\begin{subfigure}{.5\textwidth}
\centering
\includegraphics[width=\textwidth]{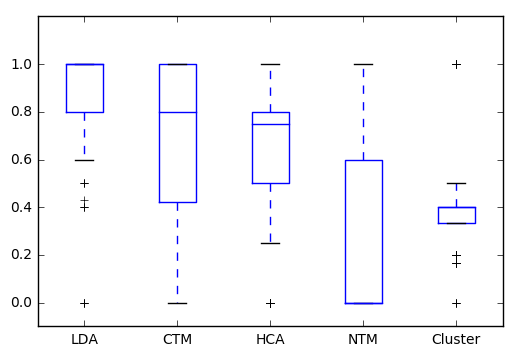}
\caption{\apnews}
\label{fig:apnews-mp}
\end{subfigure}
~
\begin{subfigure}{.5\textwidth}
\centering
\includegraphics[width=\textwidth]{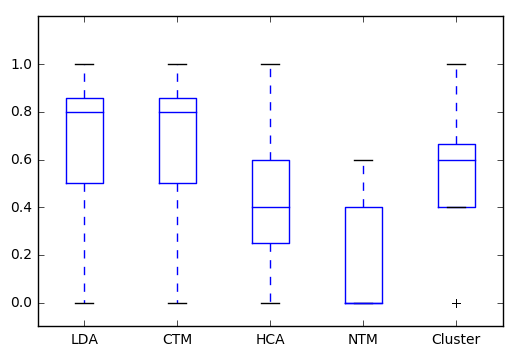}
\caption{\bnc}
\label{fig:bnc-mp}
\end{subfigure}
~
\caption{Boxplots of model precision}
\label{fig:mp-boxplots}
\end{figure*}

\begin{figure*}[t]
\begin{subfigure}{.5\textwidth}
\centering
\includegraphics[width=\textwidth]{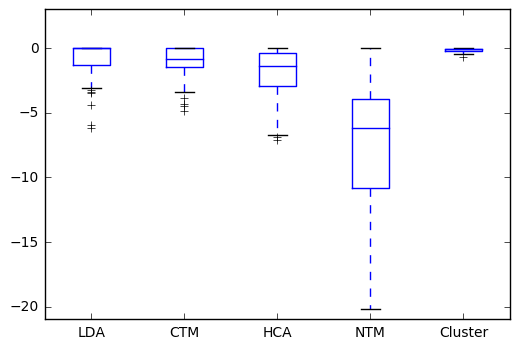}
\caption{\apnews}
\label{fig:apnews-tlo}
\end{subfigure}
~
\begin{subfigure}{.5\textwidth}
\centering
\includegraphics[width=\textwidth]{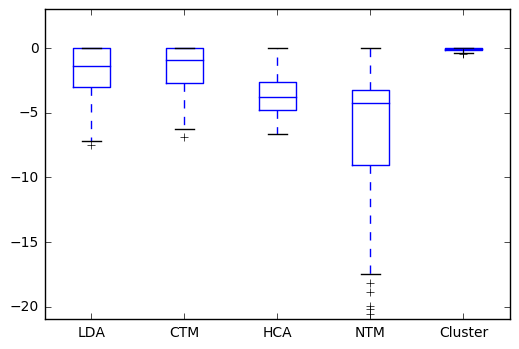}
\caption{\bnc}
\label{fig:bnc-tlo}
\end{subfigure}
~
\caption{Boxplots of topic log odds}
\label{fig:tlo-boxplots}
\end{figure*}

For our annotation task, we randomly sampled 100 documents from each of our two datasets, for each of which we generate document--topic items based on the five different topic models. In total, therefore, we annotated 1000 (100 documents $\times$ 2 collections $\times$ 5 topic models) document--topic combinations. After quality control, the final dataset contains an average of 5.4 and 5.5 valid intruder topic annotations for \apnews and \bnc, respectively. 

\newcite{Chang+:2009} proposed topic log odds (``TLO'') as a means of
evaluating the topic intrusion task. The authors defined topic log odds
for a document--topic pair as the difference in the log-probability
assigned to the intruder and the log-probability assigned to the topic
chosen by a given annotator, which they then averaged across annotators
to get a TLO score for a single document. Separately,
\newcite{Chang+:2009} proposed model precision as a means of evaluating
the word intrusion task, whereby they simply calculated the proportion
of annotators who correctly selected the intruder word for a given
topic. In addition to presenting results based on TLO, we apply the
model precision methodology in our evaluation of the topic intrusion
task, in calculating the proportion of annotators who correctly selected
the intruder topic for a given document, which we then average across
documents to derive a model score.

The results of the human annotation task are summarised in
\tabref[s]{mean-model-precision} and \ref{tab:mean-tlo}.  Looking at
model precision for \apnews first, we see that \lda outperforms the
other topic models. \ctm and \hca perform credibly, whereas \ntm and
\cluster are quite poor. Moving on to \bnc, we see a drop in score for
\lda, to a level comparable with \ctm. \cluster improves slightly higher
than \bnc, whereas \hca drops considerably (despite being designed
specifically to deal with word burstiness in the longer documents
characteristic of \bnc). \figref{mp-boxplots} shows boxplots for
topic-level model precision, and reflects a similar trend.

Looking next to TLO in \tabref{mean-tlo}, we see a totally different
picture, with \cluster being rated as the best topic model by a clear
margin. This exposes a flaw in the TLO formulation, in the case of
adversarial topic models such as \cluster which assign near-uniform
probabilities across all topics. This results in the difference in
probability mass being very close to the upper bound of zero in all
cases, meaning that even for random topic selection, TLO is near
perfect. We can also see this in \figref{tlo-boxplots}, where the boxes
for \cluster have nearly zero range. Indeed, if we combined the results
for TLO with those for topic coherence, we would (very wrongly!)
conclude that \cluster performs best over both document
collections. More encouragingly, for the other four topic models, the
results for TLO are much more consistent with those based on model
precision.




\subsection{Direct Annotation of Topic Assignments}
\label{sec:direct-app}

\newcite{newman+:2010} proposed a more direct approach to topic
coherence, by asking people to rate topics directly based on the top-$N$
words. Taking inspiration from their methodology, we propose to directly
annotate each topic assigned to a target document. We present the human
annotators with the target document and the top-ranked (highest
probability) topic from each of the five topic models, and ask them to
rate each topic on an ordinal scale of 0--3. At the model level, we take
the mean rating over all document--topic pairings for that topic model
(based, once again, on 100 documents per collection).\footnote{The 100
  documents used for this task were different to the ones used in
  \secref{topic-intrusion}.} We summarise the findings in
\tabref{obs-coherence}.

We observe that, in the case of \apnews, \lda does considerably better
than \ctm and \hca, whereas for \bnc, \lda and \ctm are quite close,
with \hca close behind. \cluster and \ntm do poorly across both
datasets. The overall trend for \apnews of \lda $>$ \ctm $>$ \hca $>$
\cluster $>$ \ntm is consistent with the model precision results in
\tabref{mean-model-precision}.  In the case of \bnc, the observation of
\ctm $\approx$ \lda $>$ \hca $>$ \cluster $>$ \ntm is also broadly the
same, except that \hca does not do as well over the topic intrusion
task. Here, we are more interested in the relative performance of topic
models than absolute numbers, although the low absolute scores are an
indication that it is a difficult annotation task.

Broadly combined across the two evaluation methodologies, \lda and \ctm 
are top-performing, \hca gets mixed results, and \cluster and \ntm are 
the lowest performers. These results generally agree with the model 
precision findings, demonstrating that model precision is a more robust 
metric than TLO.

\begin{table}[t]
\centering

\begin{tabular}{ccc}
\toprule
\multirow{2}{*}{Topic Model} & \multicolumn{2}{c}{Average rating} \\
  \cmidrule{2-3}
                             & \apnews            & \bnc            \\
       \midrule
\lda                          & 1.26              & 1.01           \\
\ctm                          & 0.96              & 1.02           \\
\hca                          & 0.95              & 0.90           \\
\ntm                          & 0.36              & 0.46           \\
\cluster                      & 0.41              & 0.66           \\
\bottomrule
\end{tabular}
\caption{Top-1 document--topic rating for each topic model}
\label{tab:obs-coherence}
\end{table}

\section{Automatic Evaluation}
\label{sec:auto-eval}

A limitation of the topic intrusion task is that it requires manual
annotation, making it ill-suited for large-scale or automatic
evaluation. We present the first attempt to automate the prediction of
the intruder topic, with the aim of developing an approach to topic
model evaluation which complements topic coherence (as motivated in
\secref[s]{topic-coherence} and \ref{sec:human-judge}).

\subsection{Methodology}
\label{sec:methodology}

We build a support vector regression (SVR) model \cite{joachims:2006} to 
rank topics given a document to select the intruder topic.  We first 
explain an intuition of the features that are driving the SVR.

To rank topics for a document, we need to first compute the probability 
of a topic $t$ given document $d$, i.e.\ $P(t|d)$.  We can invert the 
condition using Bayes rule:
\begin{align*}
P(t|d) &= \frac{P(d|t) P(t)}{P(d)} \\
    &\propto P(d|t)P(t)
\end{align*}
We can omit $P(d)$ as the probability of document $d$ is constant for 
the topics that we are ranking.

Next we represent topic $t$ using its top-$N$ highest probability words, 
giving:
\begin{align*}
 P(t|d) &\propto P(d | w_{1},...,w_{N}) P(w_{1},...,w_{N}) \\
  &\propto \log P(d | w_{1},...,w_{N}) + \\
  &\phantom{\propto \log}\log P(w_{1},...,w_{N})
\end{align*}

\begin{figure*}[t]
\begin{subfigure}{.5\textwidth}
\centering
\includegraphics[width=\textwidth]{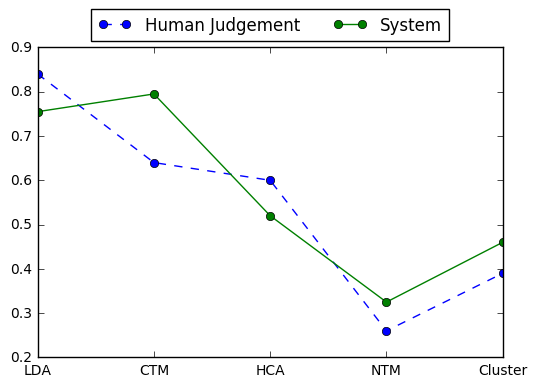}
\caption{\apnews}
\label{fig:mp-apnews-comp}
\end{subfigure}
~
\begin{subfigure}{.5\textwidth}
\centering
\includegraphics[width=\textwidth]{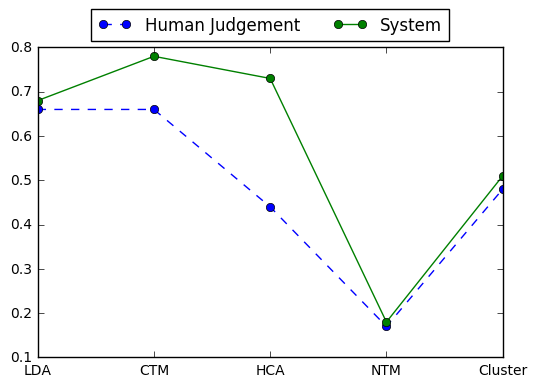}
\caption{\bnc}
\label{fig:mp-bnc-comp}
\end{subfigure}
~
\caption{Mean Model Precision Comparison}
\label{fig:mp-comparison}
\end{figure*}

 

The first term $\log P(d | w_{1},...,w_{N})$ can be interpreted from an 
information retrieval perspective, where we are computing the relevance 
of document $d$ given query terms $w_1$, $w_2$, ..., $w_N$. This term 
constitutes the first feature for the SVR. We use 
Indri\footnote{\url{http://www.lemurproject.org}} to index the document 
collection, and compute $\log P(d | w_{1},...,w_{N})$ given a set of 
query words and a document.\footnote{$N =10$.}

We estimate the second term, $\log P(w_{1},...,w_{N})$, using the 
pairwise probability of the topic words:
\begin{equation*}
\sum_{0< i \leq m}\sum_{i+1\leq j \leq m} \log 
\frac{\#(w_i,w_j)}{\#(\cdot)}
\end{equation*}
where $m$ denotes the number of topic words used, $\#(w_i,w_j)$ is the
number of documents where word $w_i$ and $w_j$ co-occur, and $\#(\cdot)$
is the total number of documents.  We explore using two values of $m$
here: 5 and 10.\footnote{That is, if $m=5$, we compute pairwise
  probabilities using the top-$5$ topic words.} These two values
constitute the second and third features of the SVR.

To train the SVR, we sample 1700 random documents and split them into 
1600/100 documents for the training and test partitions, respectively.  
The test documents are the same 100 documents that were previously used 
for intruder topics (\secref{topic-intrusion}). As the intruder topics 
are artificially generated, we can sample additional documents to create 
a larger training set for the SVR; the ability to generate arbitrary 
training data is a strength of our method.

We pool together all 5 topic models when training the SVR, thereby
generating 8000 training and 500 development and testing instances for
each dataset. For each document, the SVR is trained to rank the topics
in terms of their likelihood of being an intruder topic.\footnote{We use
  the default hyper-parameter values for the SVR ($C = 0.01$), and hence
  do no require a development set for tuning.} The top-ranking topic is
selected as the system-predicted intruder word, and model precision is
computed as before (\secref{topic-intrusion}).\footnote{Note that the
  system model precision for each document--topic combination is a
  binary value as there is only 1 system --- as opposed to multiple
  annotators --- selecting an intruder word.}

%

\subsection{System results}
\label{sec:sys-results}

In \figref{mp-comparison}, we present the human vs.\ system mean model
precision on the test partition for each of the topic models. We see
that the trend line for the system model precision very closely
tracks that of human model precision. In general, the best systems ---
\lda and \ctm~--- and the worst systems --- \ntm and \cluster~--- are
predicted correctly. The correlation between the two is very high, at
$r = 0.88$ and $0.87$ for \apnews and \bnc, respectively. This suggests
that the automated method is a reliable means of evaluating
document-level topic model quality.



\begin{table*}[t]
\begin{center}
\begin{adjustbox}{max width=1.0\textwidth}
\begin{tabular}{p{3cm}lp{15cm}}
\toprule
\tl{20}{\textbf{Error Type:}\\High human MP\\Low system MP} &
\multirow{6}{*}{\textbf{Document}} & {more than 2,000 attendees are 
expected to attend public funeral services for former nevada gov. kenny 
guinn . a catholic mass on tuesday morning will be followed by a 
memorial reception at palace station . the two-term governor who served 
from 1999 to 2007 died thursday after falling from the roof of his las 
vegas home while making repairs . he was 73 . guinn 's former chief of 
staff pete ernaut says attendance to the services will be limited only 
by the size of the venues . services start at 10 a.m. at st. joseph , 
husband of mary roman catholic church ...}  \\
\cdashline{2-3}
& \multirow{4}{*}{\textbf{Topics}} & \hla{0: church gay marriage 
religious catholic same-sex couples pastor members bishop} \\
&& 1: died family funeral honor memorial father death wife cemetery son 
\\
&& 2: casino las vegas nevada gambling casinos ford vehicles cars car \\
&& X: students college student campus education tuition universities 
colleges high degree \\
\cline{2-3}
& \multirow{6}{*}{\textbf{Document}} & {the milwaukee art museum is 
exhibiting more than 70 works done by 19th century portrait painter 
thomas sully . it 's the first retrospective of the artist in 30 years 
and the first to present the artist 's portraits and subject pictures .  
sully was known for employing drama and theatricality to his works . in 
some of his full-length portraits , he composed his figures as if they 
were onstage . some of his subjects even seem to be trying to directly 
engage the viewer . milwaukee art museum director daniel keegan says the 
exhibit provides a new look ...}  \\
\cdashline{2-3}
& \multirow{4}{*}{\textbf{Topics}} & 0: china art chinese arts artist 
painting artists cuba world beijing \\
&& \hla{1: show music film movie won festival tickets game band play} \\
&& 2: online information internet book video media facebook phone 
computer technology
 \\
&& X: kelley family letter leave absence left united jay weeks director
\\

\midrule
\tl{20}{\textbf{Error Type:}\\Low human MP\\High system MP} &
\multirow{6}{*}{\textbf{Document}} & {( ap ) ? the west virginia lottery 
is celebrating its 28th birthday by doing what it does best : awarding 
large sums of money . the lottery will mark the milestone on thursday by 
giving away prizes of \$ 1 million , \$ 100,000 and \$ 10,000 . the 
three finalists were selected out of thousands of entries from the 
lottery 's monopoly millionaire instant game . the finalists are josh 
schoolcraft of given , douglas schafer of wheeling and todd kingrey of 
charleston .  all three are due at lottery headquarters in charleston to 
collect their winnings ...
}  \\
\cdashline{2-3}
& \multirow{4}{*}{\textbf{Topics}} & 0: jackpot powerball mega lottery 
lotto jackpots prizes ticket megaplier tickets \\
&& 1: mingo earl wheeling virginia ap charleston wvu huntington 
coalfields rockefeller
 \\
&& \hlb{2: museum artifacts exhibit paintings artwork historical curator 
sculpture exhibition exhibits} \\
&& X: abercrombie ridley solace daley enclosures hobbyists hawaiian 
seventeen secondhand probate \\
\cline{2-3}
& \multirow{6}{*}{\textbf{Document}} & {a 75-year-old driver has died 
after a collision near o'neill in northern nebraska . the holt county 
sheriff 's office says the accident occurred wednesday afternoon , less 
than a mile east of o'neill . the office says thomas schneider halted at 
a stop sign and then turned east onto nebraska highway 108 . but he 
apparently turned too wide and went into the oncoming lane . his vehicle 
struck a westbound vehicle driven by 52-year-old gerald kemp , of 
niobrara . schneider was pronounced at the scene . the sheriff 's office 
says kemp suffered no visible injuries ...}  \\
\cdashline{2-3}
& \multirow{4}{*}{\textbf{Topics}} & 0: officers shot car shooting 
officer sheriff woman died killed hospital
 \\
&& \hlb{1: service weather area storm miles airport snow river bridge 
emergency} \\
&& \hlb{2: prison prosecutors charges guilty trial judge case charged 
murder pleaded}
\\
&& X: toll road rocky carpenter hogan indiana long harvey private 
director
\\
\bottomrule
\end{tabular}
\end{adjustbox}
\end{center}
\caption{Document and topic examples for two types of errors. ``MP'' 
  denotes model precision, ``X'' the intruder topic, and the indices the
  ranking of the topics. 
  Topics highlighted in pink (yellow) are those incorrectly selected by the system (humans) 
  as intruder topics.}
\label{tab:doc-example}
\end{table*}

\begin{table*}[t!]
\begin{center}
\begin{adjustbox}{max width=1.0\textwidth}
\begin{tabular}{p{2.3cm}lp{15cm}}

\toprule
\tl{8}{\lda} &
\multirow{6}{*}{\textbf{Document}} & {more than 2,000 attendees are 
expected to attend public funeral services for former nevada gov. kenny 
guinn . a catholic mass on tuesday morning will be followed by a 
memorial reception at palace station . the two-term governor who served 
from 1999 to 2007 died thursday after falling from the roof of his las 
vegas home while making repairs . he was 73 . guinn 's former chief of 
staff pete ernaut says attendance to the services will be limited only 
by the size of the venues . services start at 10 a.m. at st. joseph , 
husband of mary roman catholic church ...
}  \\
\cdashline{2-3}
& \multirow{2}{*}{\textbf{Topics}} & 0: church gay marriage 
religious catholic same-sex couples pastor members bishop \\
&& 1: died family funeral honor memorial father death wife cemetery son \\

\midrule

\tl{9}{\hca} &
\multirow{6}{*}{\textbf{Document}} & {usa today founder al neuharth has died in cocoa beach , florida . he was 89 . the news was announced friday by usa today and by the newseum , which he also founded . neuharth changed american newspapers by putting easy-to-read articles and bright graphics in his national daily publication , which he began in 1982 when he ran the gannett co. newspaper group . he wanted to create a bright , breezy , fun newspaper that would catch people 's attention and not take itself too seriously. its annual revenues increased from 200 million to more than 3 billion ...
}  \\
\cdashline{2-3}
& \multirow{3}{*}{\textbf{Topics}} & 0: honorary commencement philanthropist journalism distinguished honored bachelor pulitzer doctorate harvard
\\
&& 1: shortfall premiums budget reductions cuts shortfalls salaries pensions revenues budgets \\

\midrule

\tl{8}{\ctm} &
\multirow{6}{*}{\textbf{Document}} & {a teenage driver who survived a southeastern indiana crash that killed three other youths will spend 90 days in juvenile detention and surrender his driver 's license until age 21 . the 17-year-old driver admitted to charges of reckless homicide and reckless driving during a ripley county juvenile court hearing thursday in versailles , indiana state police sgt. noel houze jr. told the associated press . the teenager choked back sobs throughout the half-hour hearing . the teen will be sent to a juvenile facility in muncie . he also must complete 350 hours of community service ..
}  \\
\cdashline{2-3}
& \multirow{2}{*}{\textbf{Topics}} & 0: officers shot car shooting officer sheriff woman died killed hospital \\
&& 1: prison prosecutors charges guilty trial judge case charged murder pleaded  \\

\midrule

\tl{9}{\ntm} &
\multirow{6}{*}{\textbf{Document}} & {a judge in will county has approved further testing on the coat an oswego man was wearing when his wife and three children were found shot to death in 2007 . christopher vaughn is accused of killing his family inside their suv , which was parked on a frontage road along interstate 55 . authorities found kimberly vaughn shot to death , along with their children , 12-year-old abigayle , 11-year-old cassandra and 8-year-old blake . assistant state 's attorney mike fitzgerald on monday said prosecutors asked for more dna testing on the coat ...
}  \\
\cdashline{2-3}
& \multirow{3}{*}{\textbf{Topics}} & 0: arraigned burglarizing arrested bigamy detectives motorcyclist arraignment coroner accomplice fondled \\
&& 1: quarterly pretax dividend profit annualized earnings profits stockholders writedown premarket  \\

\midrule
\tl{8}{\cluster} &
\multirow{5}{*}{\textbf{Document}} & {a southwest idaho district court judge has been arrested on suspicion of misdemeanor driving under the influence . the idaho press-tribune reports ( http://bit.ly/npiita ) that 3rd district court judge renae hoff was taken into custody early saturday morning in meridian . meridian deputy police chief tracy basterrechea says an officer pulled the 61-year-old hoff over after she failed to " maintain the lane of travel . "
}  \\
\cdashline{2-3}
& \multirow{3}{*}{\textbf{Topics}} & 0: suppliers manufacturers companies importers supplier exporters distributors market wholesalers export  \\
&& 1: deported deportation incarcerated prison detention jail parole imprisoned convicts incarceration \\

\bottomrule
\end{tabular}
\end{adjustbox}
\end{center}
\caption{Example documents and their corresponding topics for different topic models}
\label{tab:doc-topic-examples}
\end{table*}

\section{Discussion}
\label{sec:discussion}


To better understand the differences between human- and system-predicted
intruder topics, we present a number of documents and their associated
topics in \tabref{doc-example}, focusing specifically on: (a) intruder
topics that humans struggle to identify but our automatic method
reliably detects; and (b) conversely, intruder topics which humans
readily identify but our method struggles to detect. 

Looking at the topics across the two types of errors, we notice that
there are often multiple ``bad'' topics for these documents:
occasionally the annotators are able to single out the worst topic while
the system fails (1st and 2nd document), but sometimes the opposite
happens (3rd and 4th document).  In the first case, the top-ranking
topic (\ex{church, gay, ...}) from the topic model is associated with
the document because of the service, but actually capturing a very
different aspect of religion to what is discussed in the document, which
leads our method astray. A similar effect is seen with the second
document. In the case of the third and fourth documents, there is
actually content further down in the document which is relevant to the
topics the human annotators select, but it is not apparent in the
document snippet presented to the annotators. That is, the effect is
caused by resource limitations for the annotation task, that our
automated method does not suffer from.

When we aggregate the top-level model precision values for a topic
model, these differences are averaged out (hence the strong correlation
in \secref{sys-results}), but these qualitative analyses reveal that
there are still slight disparities between human annotators and the
automated method in intruder topic selection.

To further understand how the topics relate to the documents in different topic models, we present 
documents with the corresponding topics for different topic models in \tabref{doc-topic-examples}.

In the human annotation task, we use the top-10 most probable words to
represent a topic. We use 10 words as it is the standard approach to
visualising topics, but this is an important hyper-parameter which needs
to be investigated further \cite{lau+:2016}, which we leave to future work.



\section{Conclusion}
\label{sec:conclusion}

We demonstrate empirically that there can be large discrepancies between
topic coherence and document--topic associations. By way of designing an
artificial topic model, we showed that a topic model can simultaneously
produce topics that are coherent but be largely undescriptive of the
document collection.  We propose a method to automatically predict
document-level topic quality and found encouraging correlation with
manual evaluation, suggesting that it can be used as an alternative
approach for extrinsic topic model evaluation.


\section*{Acknowledgements}

This research was supported in part by the Australian Research Council.

\bibliographystyle{emnlp_natbib}
\bibliography{emnlp2017,papers}

\end{document}